\documentclass[11pt,a4paper]{article}
\usepackage[hyperref]{acl2020}
\usepackage{times}
\usepackage{pbox}
\usepackage{xcolor}
\usepackage{latexsym}

\usepackage{fourier} 
\usepackage{array}
\usepackage{makecell}

\usepackage{booktabs}
\newcommand{\tabitem}{~~\llap{\textbullet}~~}

\usepackage{microtype}
\usepackage{linguex}
\usepackage{amsthm}
\usepackage{graphicx}

\theoremstyle{definition}
\newtheorem{exmp}{Example}[section]

\usepackage{array}

\usepackage{diagbox}
\aclfinalcopy 

\title{Who is \textit{we}? Disambiguating the referents of first person plural pronouns\\ in parliamentary debates}

\author{Ines Rehbein \\
  Data and Web Science Group \\
  University of Mannheim \\
  \\
  \hspace{2.5cm}\texttt{ines@informatik.uni-mannheim.de} \\\And
  Josef Ruppenhofer \\
  Archive for Spoken German \\
  Leibniz Institute \\
  for the German Language \\
  \hspace{7.5cm}\texttt{ruppenhofer@ids-mannheim.de}\\\And
  Julian Bernauer \\
   MZES \\
  University of Mannheim \\
    \\
    \\
 \hspace{-3.8cm} \texttt{julian.bernauer@mzes.uni-mannheim.de} \\}

\date{}

\begin{document}
\maketitle
\begin{abstract}
This paper investigates the use of first person plural pronouns as a rhetorical device in political speeches. We present an annotation schema for disambiguating pronoun references and use our schema to create an annotated corpus of debates from the German Bundestag. We then use our corpus to learn to {\em automatically} resolve pronoun referents in parliamentary debates. We explore the use of data augmentation with weak supervision to further expand our corpus and report preliminary results.
\end{abstract}

\section{Introduction}
\label{sec:introduction}

Personal pronouns are an important rhetorical device in political speeches that allow politicians to shape their message to appeal to specific audiences.
Multiple functions of pronouns have been described, such as creating a feeling of unity with the audience (\ref{ex:unity}), sharing responsibility (\ref{ex:responsibility}) or criticising others (\ref{ex:criticism}) \cite{beard:2000,bramley:2001,hakansson:2012}.

\begin{exmp}
 Members of Congress, we must work together to help control those costs (Bush 2004)
 \label{ex:unity}
\end{exmp}

\begin{exmp}
 We have increased our budget at a responsible 4 percent (Bush 2001)
 \label{ex:responsibility}
\end{exmp}

\begin{exmp}
the more we get involved with other people, the more complicated our relationships get (B. Clinton 2002)
 \label{ex:criticism}\footnote{Two of the examples taken from \newcite{hakansson:2012}.}
\end{exmp}

\newcite{tyrkko:2016} calls personal pronouns ``one of the primary linguistic features used by political speakers to manage their audiences' perceptions of in-groups and out-groups''. 
This makes them especially important for populist rhetoric where the speaker evokes a dichotomous  view of society, {\em us-versus-them} (see, e.g., \newcite{mudde:2004,mudde:kaltwasser:2017}).

While the practice of {\em othering}  might seem to be the most prominent feature of personal pronouns in political discourse,  
another important aspect also needs to be considered, namely their referential ambiguity \cite{tyrkko:2016,wales:1996}.
As stated by \newcite[pp.12]{allen:2007}, 
\begin{quotation}
``Shifting identity through pronoun choice and using pronouns with ambiguous referents enables politicians to appeal to diverse audiences which helps broaden their ability to persuade the audience to their point of view. It is a scattergun effect \textemdash shoot broadly enough and you'll hit something''.  
\end{quotation}

While prior research on the interface between corpus linguistics, pragmatics, discourse studies and political science has presented empirical findings based on word frequencies \cite{vukovic:2012,tyrkko:2016,alavidze:2017}, only few studies have tried to systematically investigate this topic in more detail, i.e., by trying to measure the agreement between human annotators for disambiguating the referents of personal pronouns in political speeches, or by presenting large-scale studies of the use of personal pronouns beyond word frequencies.

This paper takes first steps in that direction by means of an annotation study in which we classify instancse of the first person plural pronoun \textit{wir} `we' in German parliamentary debates, using a classification scheme with 9 different classes. We report inter-annotator agreement for this highly subjective task and analyse our disagreements. We then present a preliminary analysis of our data where we look into  differences in the use of {\em we/us} in political speeches, depending on (i) the speaker, (ii) the topic, and (iii) the speaker's party affiliation.

In the second part of the paper, we undertake first experiments towards automatically predicting the referents of first person pronouns in parliamentary debates. For that, we make use of transfer learning, in combination with data augmentation based on weak supervision \cite{snorkel:2016,snorkel:2020}. We show that our transfer learning approach brings substantial improvements over a majority baseline while pretraining the model on the larger, noisy data and fine-tuning it on our manual annotations yields only small improvements over training on the manual annotations only.

\section{Related Work}
\label{sec:relwork}

\paragraph{First person plural pronouns from a linguistic perspective}

The reference of German \textit{wir}, just like that of English \textit{we}, is quite variable. Following the typology of \newcite{cysouw2002impact}, German \textit{wir} as a first person plural (1PL) form has multiple distinct uses: (i) \texttt{minimal inclusive}, consisting of speaker and hearer (\ref{ex:minimincl}); (ii) \texttt{augmented inclusive}, adding third parties beyond the minimal inclusive (\ref{ex:augincl}); (iii) \texttt{exclusive}, consisting of the speaker and third parties, but excluding the hearer (\ref{ex:excl}). 

\begin{exmp}
\label{ex:minimincl} Sollen wir morgen telefonieren?\\
`Shall we talk on the phone tomorrow?'
\end{exmp}
\begin{exmp}
\label{ex:augincl} Kim kommt um 12 an.  Sollen wir dann Mittag essen gehen?\\
`Kim will arrive at 11. Shall we go to lunch then?' [all three of us]
\end{exmp}
\begin{exmp}
\label{ex:excl} Wir gehen ins Kino. Was habt ihr vor?\\
`We're going to the movies. What are your plans?'
\end{exmp}

In addition, special subtypes of uses may be recognized. For English, \newcite{Quirketal1985} discuss a set of special (sub)uses that also occur with German \textit{wir}. For instance, a single author may nevertheless use 1PL pronouns to avoid appearing `egotistical'. Doctors (among others) may use the 1PL pronoun in a a hearer-oriented way (e.g. \textit{How are we feeling today?}). Of greatest relevance to our data are \newcite{Quirketal1985}'s generic uses and their class of rhetorical uses where the pronoun refers to a collective such as `the party', `the nation'.

While linguistic analyses of pronouns often simply view them as words with determinate reference to a deictically, anaphorically or cataphorically available entity, pragmatic and discourse-oriented studies of pronouns like ours focus on their conceptual emptiness and the fact that their referents must be inferred in context, with the possibility of (un)intentionally ambiguous uses, since individuals have multiple social, discursive and interactional roles.

\paragraph{Corpus studies of 1PL reference}

Very close in spirit to our work but operating on conversational interactions and with  categories appropriate to that domain, \newcite{scheibman2014referentiality} presents a study on the reference of \textit{we} in relation to predicate patterns and pragmatic functions.
The study coded instances of \textit{we} from the Santa Barbara Corpus of Spoken American English for several features, among them (i) the inclusive vs exclusive distinction, (ii) type of referent (e.g. family member, couple, classmates, human beings, etc.), (iii) tense of predicate, (iv) modals  present.  
The authors' findings suggest that different referential uses of first personal pronouns may be distinguishable based on contextual cues such as tense and modality.

\paragraph{Pronouns in political discourse} \newcite{tyrkko:2016} presents a diachronic study of the use of personal pronouns in political speeches over two centuries, showing shifts from a self-centric style (marked by frequent use of \textit{I}) towards the more inclusive use of 1PL forms in the 1920s, which the author ties to the emergence of broadcast media. The study does not disambiguate 1PL forms but counts all of them as inclusive. 

\newcite{inigo2004use} studies the use of \textit{we} in 5 Question Time Sessions of the British parliament, where MPs ask questions of government ministers. She distinguishes what she calls exclusive, inclusive, generic and parliamentary uses of \textit{we} and examines their distribution across different combinations of interactants (opposition MP to member of government; member of government to opposition MP; member of government and supportive MP (in either direction)).\footnote{There is no generally agreed-upon terminology used to distinguish uses of \textit{we}, either in general or in the political or parliamentary context. For Inigo-Mora the generic \textit{we} refers to "a kind of patriotic "we" that embraces all British people". In the terminology of \newcite{Quirketal1985} this would be called a collective use. In our annotation scheme, the uses at issue would be labeled "COUNTRY".} The frequency distribution is interpreted along two dimensions: (i) power and distance and (ii) identity, community and persuasion. Among the findings is that exclusive uses of \textit{we} constitue the most common type overall, accounting for 53.4\% of all tokens. Exclusive \textit{we} is at its most dominant in interactions from government supporting MPs to opposition MPs (76.1\%) while it is hardly ever used in questions from opposition MPs to a member of government, which is taken to reflect the power dynamics. Inclusive uses of \textit{we} were found to be much rarer overall, making up 14.5\% of all tokens. None of these are uttered by opposition members speaking to members of government, while three quarters are produced between government supporting MPs and members of government, expressing shared identity. Opposition MPs mostly use generic and parliamentary \textit{we}, thus affiliating themselves with the parliament as a distinct branch of government and the country at large, likely because that is where persuasion is most likely to succeed. It is unclear to what extent these results carry over to the plenary setting. 

\paragraph{Non-parliamentary political discourse} Studies of 1PL pronouns have also targeted other types of interactions. \newcite{bullfetzer2006} analyze the use of \textit{you} and \textit{we} in tv interviews with British politicians that were broadcast during the 1997 and 2001 British general elections and just before the war with Iraq in 2003. The focus of the study was on question-response sequences in which politicians make use of pronominal shifts as a means of equivocation to effect shifts of accountability and responsibility.  
\newcite{PROCTOR20113251} examine the use of \textit{we} by four (vice-)presidential candidates in debates and interviews around the time of the 2008 US election. The study focuses on which groups are the referents of \textit{we} and which entities are picked out by possessive NPs of the form \textit{our N}, considering the results in light of the candidates' political stature and targeted office as well as the differences between debate and interview settings. 
\paragraph{Politeness} Finally, we note that quite a lot of research on pronoun use exists in the area of politeness, though this typically targets pronouns of address. For instance, in a seminal study, \newcite{Brown-and-Gilman-1960} discussed the differences in use between informal and formal second person pronouns (such as German \textit{du} and \textit{Sie}) as forms of address in terms of their association with the dimensions of power and solidarity between speakers. The authors argue that, while for a long time the form chosen was mainly determined by power differentials, over time the choice came to depend more on the factor of solidarity.

\section{Annotation Study} 

\begin{table}[t]\small
 \begin{center}
  \begin{tabular}{|l|r|r|r|r|}
    \hline
{\bf Party} & {\bf \#Tokens} &  {\bf \#Annot} & {\bf \#Spk} & {\bf per 1000} \\
    \hline
AfD 	    &  8,993  &	 142 & 8 &  15.8 \\
CDU/CSU     & 10,674  &	 335 & 5 &  31.4 \\
FDP 	    &  7,358  &	 166 & 7 &  22.6 \\
GRÜNE       &  7,457  &	 136 & 5 &  18.2 \\
LINKE       &  9,310  &	 130 & 6 &  14.0 \\
SPD 	    &  7,438  &	 245 & 4 &  32.9 \\
fraktionslos&    797  &	   9 & 1 &  11.3 \\ 
\hline
{\bf Total} & 52,027 & 1,163 & 36 &  22.3 \\ 
\hline
\end{tabular}

\end{center}
\caption{Some statistics for the annotated testset ({\bf \#Spk}: no. of speakers per party; {\bf per 1000}: no. of 1PL pronouns per 1000 tokens).}\label{tab:testset}
\end{table}

\begin{table*}[t]
 \begin{center}\small
  \begin{tabular}{|l|l|l|}
   \hline
   {\bf Class}   &   {\bf Description} &   {\bf Example} \\
   \hline
   {\sc Board}   & Members of a board/& {\bf Wir}  haben heute im   \\
                 & commission/committee & Untersuchungsausschuss erfahren \\
   \hline
   {\sc Country} & references to Germany/ &  {\bf Wir} sind Weltmeister \\
                 & all Germans               &  {\bf Unser} Grundgesetz \\
   \hline
   {\sc Generic} & generic uses that can be replaced & Daran werden {\bf wir} {\bf uns}  \\
                 & by {\em one (de: man)}            & noch in 100 Jahren erinnern \\
   \hline
   {\sc Govern}  & members of the government & {\bf Wir} haben die Arbeitslosigkeit bek\"{a}mpft.  \\
   \hline
   {\sc Parl}    & members of the parliament  & {\bf Wir} Abgeordnete... \\
                 &                            & Lassen Sie {\bf uns} diesen Antrag heute beschlie{\ss}en \\
   \hline
   {\sc Party}   & members of one specific party &  {\bf Wir} Liberale haben schon fr\"{u}her...\\ 
   \hline
   {\sc People}  & groups of people defined by social & Wie {\bf wir} Älteren {\bf uns} verhalten... \\
                 & variables (age, profession, religion & {\bf Wir} Steuerzahler, {\bf Wir} Christen,  \\ 
                 & and other shared characteristics ...) & {\bf Wir} Pendler, ... \\
   \hline
   {\sc SpecPers}& groups of individuals or     & {\bf Wir} beide haben dar\"{u}ber diskutiert \\

                 & members of more than one group  & {\bf Wir}, die deutsche und die israelische Regierung  \\
   \hline
   {\sc Union}   &  geo-political groups on a &   {\bf Wir} in der EU... \\
                 &  supranational level (EU, NATO) &  {\bf Unsere} Europ\"{a}ische Union... \\
   \hline
  \end{tabular}
 \end{center}
 \vspace{-0.2cm}
\caption{Overview of the annotation scheme for 1PL references in parliamentary debates.}\label{tab:schema}
 \vspace{-0.2cm}
\end{table*}

\subsection{Data}
\label{sub:data}

The data we use in our study are parliamentary debates from the German Bundestag, covering a time period from Oct 24, 2017 to May 19, 2021.\footnote{The data is available in XML format from \url{https://www.bundestag.de/services/opendata}.} 
The corpus includes over 330,000 sentences ($>$16,5 mio tokens), with political speeches by 777 different speakers.

From the XML files, we extracted the individual speeches and randomly selected a subset for manual annotation where we tried to collect roughly the same number of speeches/tokens for each party (see table \ref{tab:testset}).
This resulted in a testset with 36 speeches by different speakers (52,027 tokens) 
 where we manually disambiguated all instances of first person plural pronouns ({\em wir, uns, unser, unsere, unseren, unseres, unsre}) by classifying them into nine predefined classes.
 We describe our annotation schema below (\S\ref{sub:schema}).

\subsection{Annotation schema}
\label{sub:schema} 

Table \ref{tab:schema} and Table \ref{tab:schema-long} in the appendix give an overview over our classification schema.
We assume that 
references of {\em we/us} in parliamentary debates can be assigned to a small number of different categories, such as ``we, the {\sc Parliament}'' or ``our {\sc Country}'', or ``our political {\sc Party}''.
The schema has been designed in a bottom-up, data-driven fashion, using speeches from the European parliament and the German Bundestag for schema development.
We test our classification schema in an annotation experiment and investigate a) how well human annotators agree when disambiguating 1PL pronouns in political speech; b) whether it is possible to automatically predict the intended reference of personal pronouns in parliamentary debates.

We expect that, as noted in section \ref{sec:relwork}, a large part of vagueness and ambiguity in political speech is intended and will result in low IAA between some of the classes in our classification schema. However, we also expect that some classes (such as {\sc Party}) are less ambiguous which should be reflected in a higher agreement between the annotators.

\subsection{Annotation}
\label{sub:annotation}

The annotators, two computational linguists,\footnote{The data was annotated by the first two authors of the paper.} were presented with the speech texts where all instances of 1PL pronouns were highlighted. The task then consisted in assigning a label to each pronoun.\footnote{We used INCEpTION \cite{tubiblio106270} as annotaton tool.} 
The annotators were only allowed to assign exactly one label per instance. 

\paragraph{Inter-Annotator Agreement (IAA)}  
We report Krippendorff's $\alpha$ and percentage agreement for two annotators on the 1,163 annotated instances.  
Inter-rater agreement was quite high with 0.82 $\alpha$.
%
%
Table \ref{tab:iaa}, however, shows substantial differences in agreement between the individual classes. We obtained very high agreement for {\sc Country} and {\sc Party} ($>90$\% F1) and slightly lower but still reasonably high agreement for {\sc Government, Parlament} and {\sc Union} (between $78-87$\% F1). For {\sc Generic, People} and {\sc Specific\_Persons}, agreement was substantially lower ($58-66$\% F1). Those classes are also less frequent in the data. The remaining class, {\sc Board}, was  too rare in our testset to report meaningful results  (1 instance only).\footnote{The confusion matrix for the annotations can be found in the appendix, Table \ref{tab:cm}.} We kept this class despite its low frequency in the Bundestag corpus, as we found it to be more frequent in speeches from the European Parliament.

After the annotation was completed, the two annotators discussed and resolved all disagreements to create a ground truth dataset that we used as evaluation data in our experiments (\S\ref{sec:experiments}).

\begin{table}[t]
 \begin{center}\small
  \begin{tabular}{|l|r|r|}
\hline
{\bf Class}  & {\bf F1} & {\bf Support} \\
\hline
{\sc Board}   &  0.0 &   1 \\
{\sc Country} & 92.0 & 411 \\
{\sc Generic} & 65.2 &  67 \\
{\sc Govern}  & 87.2 & 167 \\
{\sc Parl}    & 86.6 & 299 \\
{\sc Party}   & 90.6 & 103 \\
{\sc People}  & 66.7 &  13 \\
{\sc SpecPer} & 58.8 &  20 \\
{\sc Union}   & 78.2 &  82 \\ 
\hline 
Total   & 86.1 & 1,163 \\ 
\hline
\end{tabular}      
 \end{center}
\caption{IAA (F1) and support (number of annotated instances in the gold standard) for individual classes.}\label{tab:iaa}
\end{table}

\label{sec:annot}

\section{Data Analysis}
\label{sec:analysis}

We now present a preliminary analysis on our manually annotated dataset where we focus on differences in the use of 1PL pronouns across politicians and parties.

\begin{table*}[t!]
 \begin{center}\small
  \begin{tabular}{|l|r|r|r|r|r|r|r|r|r|}
    \hline
Party   & BOARD       & COUNTRY & GENERIC   & GOVERN     & PARL      & PARTY     & PEOPLE   & SPECP  &  UNION     \\ 
\hline 
AfD     & 0.0 (0) &   6.0 ~(54) &  0.6 ~(5) &  0.0 ~~(0) &  5.1 (46) &  3.4 (31) &  0.4 (4) &  0.2 (2) &  0.0 ~(0) \\ 
CDU/CSU & 0.0 (0) &  11.4 (122) &  2.1 (22) &  9.6 (102) &  5.0 (53) &  0.5 ~(5) &  0.3 (3) &  0.4 (4) &  2.2 (24) \\  
FDP     & 0.0 (0) &   5.7 ~(42) &  1.6 (12) &  0.0 ~~(0) &  6.1 (45) &  5.2 (38) &  0.0 (0) &  0.5 (4) &  3.4 (25) \\   
GRÜNE   & 0.0 (0) &   5.9 ~(44) &  1.7 (13) &  0.1 ~~(1) &  7.8 (58) &  1.2 ~(9) &  0.5 (4) &  0.7 (5) &  0.3 ~(2) \\  
LINKE   & 0.1 (1) &   7.1 ~(66) &  0.9 ~(8) &  0.0 ~~(0) &  3.7 (34) &  1.7 (16) &  0.2 (2) &  0.0 (0) &  0.3 ~(3) \\ 
SPD     & 0.0 (0) &  10.6 ~(79) &  0.9 ~(7) &  8.6 ~(64) &  8.1 (60) &  0.5 ~(4) &  0.0 (0) &  0.4 (3) &  3.8 (28) \\   
frakt.los & 0.0 (0) &   5.0 ~~(4) &  0.0 ~(0) &  0.0 ~~(0) &  3.8 ~(3) &  0.0 ~(0) &  0.0 (0) &  2.5 (2) &  0.0 ~(0) \\

\hline          
{\bf Total} & 1 & 411 & 67 & 167 & 299 & 103 & 13 & 20 & 82 \\  
\hline
\end{tabular}

 \end{center}
\caption{Distribution of classes in the annotated testset (frequency per 1000 tokens and raw counts in brackets). }\label{tab:testset2}
\end{table*}

\begin{figure*}[t!]
 \begin{center}
  \includegraphics[width=0.45\textwidth]{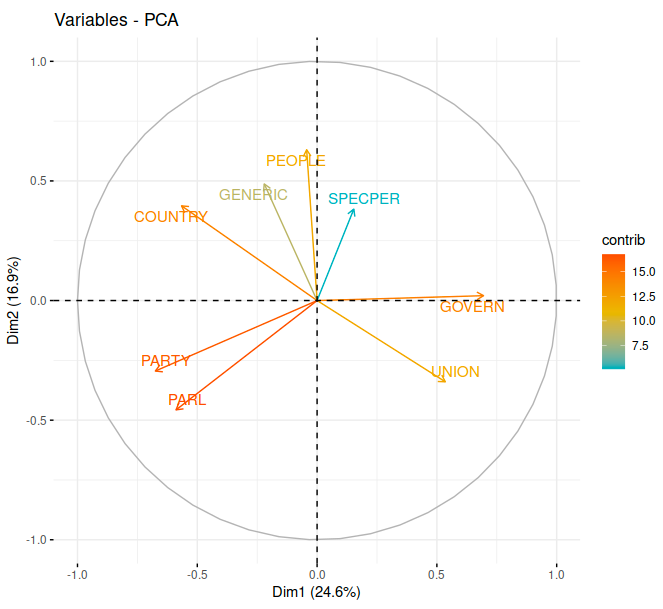} \hfill\includegraphics[width=0.45\textwidth]{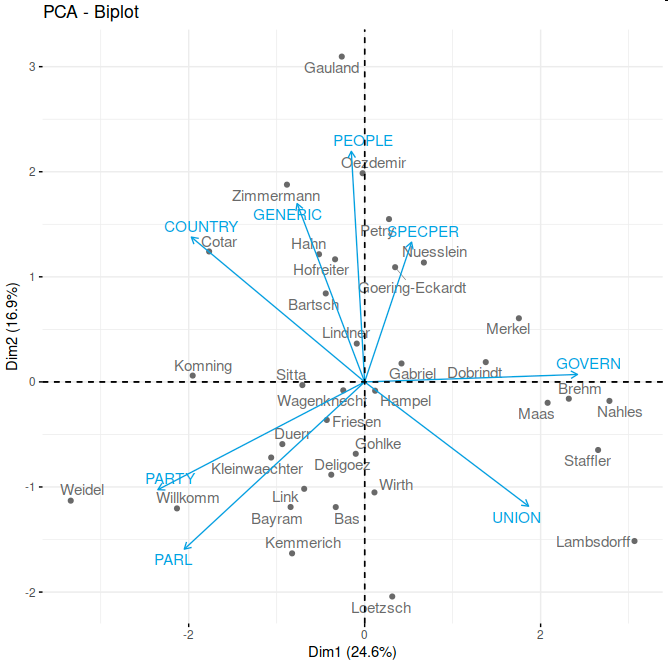}
 \end{center}
\caption{Principal Components Analysis (PCA): left figure shows the loadings for our class variables along the first two components (PC1, PC2), right figure also plots the speakers for PC1 and PC2.}\label{fig:pca}
\end{figure*}

Table \ref{tab:testset} shows that the governmental parties produce the most 1PL instances per 1000 words, which makes sense given that their members can choose between the greatest number of collective identities. 

Table \ref{tab:testset2} shows the distribution of the different classes across parties. As expected, only members of the CDU/CSU and SPD, the two parties involved in the government at the time of data collection, used {\em we} to refer to the government. Notably government MPs invoke their GOVERN identity substantially more than their PARTY identity. By contrast, members of the opposition parties refer more often to their own party, often to criticise the government and to distinguish their own policies from those of the government. This is particularly true for the FDP and the AfD, and to a lesser extent also for the LINKE and the GRÜNE. 

All parties make frequent references to the parliament ({\sc Parl}). The two parties in government, however, use many more references to {\sc Country} than the opposition parties. This observation is in contrast to the findings of \newcite{inigo2004use} (see Section \ref{sec:relwork}) who found more pronoun references to the country from members of the opposition. We would like to stress that our data is not yet large enough to produce representative results. In addition, we would also expect an impact of interaction type on the use of pronouns. \newcite{inigo2004use} investigated Question Time sessions in the British parliament while we focus on plenary speeches, which are longer,  less interactive and always have a mixed audience of supporters and opponents, whereas Question Time (superficially) addresses only one or the other. These differences might be reflected in different communicative strategies and stylistic choices.

Another reason for the higher ratio of {\sc Country} references in speeches by members of the governmental parties may be that their ranks include key office holders such as the minister of foreign affairs, whose topics tend to skew (inter)national. To investigate this, more data is needed so that we can control for the effects of office holders.

Figure \ref{fig:pca} (left) shows the loadings for our class variables along the first two dimensions of a Principal Components Analysis (PCA), based on the normalised frequency counts for the different class variables for individual speakers. The first dimension (X axis) reflects 1PL pronoun references to the government on the right-hand side and to specific parties or the parliament as a whole on the left-hand side. This opposition separates politicians from the governmental parties from the ones from the opposition parties along the first dimension (Figure \ref{fig:pca}, right). 

Figure \ref{fig:pca} (right) also seems to show topical effects as Lambsdorff, a member of the FDP and the EU parliament, is positioned closest to the vector showing the loadings of the {\sc Union} variable. This might explain why he, as the only non-governmental politician, is also positioned at the right end of the first dimension. 
The politicians that are positioned left-most on the first dimension are Weidel (AfD), Willkomm (FDP), Komning (AfD) and Cotar (AfD).  For the members of AfD, a nationalist and right-wing party deeply opposed to the European Union, it seems plausible that they are positioned not only at the opposite end of {\sc Govern} but also of {\sc Union}. Further analysis is needed to investigate this.

Figure \ref{fig:pca} (left) also shows that while the two classes {\sc Party} and {\sc Parl} are highly correlated and in opposition to {\sc Government}, the more generic classes {\sc Country, Generic} and {\sc People} also seem to cluster together. This again seems like a promising start for a more detailed analysis.
Once more data has been annotated, it will be interesting to include the topic of the speeches in the analysis. This can be easily done, either based on the agenda of the debates or by using topic models. At the moment, however, our data is still too sparse for a more fine-grained analysis.

\section{Training Data Augmentation}
\label{sec:corpus}

We now investigate whether and how well we are able to resolve ambiguities in 1PL pronoun references in parliamentary debates {\em automatically}, using our small annotated dataset to train a supervised ML system.

\begin{table}[t]
 \begin{center}\small
  \begin{tabular}{|l|r|r|r|}
 \hline   
 {\bf Class} &  {\bf \#Pattern} &  {\bf \#Hits} & {\bf err/N} \\
  \hline   
   {\sc Board}      &  1 &      7 &  0/7   \\
   {\sc Country}    &  5 &  8,795 &  2/25  \\
   {\sc Generic}    &  4 &    307 &  8/25    \\
   {\sc Govern}     &  5 & 14,851 &  2/25  \\
   {\sc Parliament}  &  4 &  3,339 &  2/25  \\
   {\sc Party}      & 11 &  8,265 &  4/25  \\
   {\sc People}     &  1 &    230 & 19/25 \\
   {\sc SpecPer}    &  4 &    106 &  3/25  \\
   {\sc Union}      &  4 &    540 &  3/25  \\ 
\hline
   {\sc Total}      &  40 & 36,433 & 43/203 \\
\hline
   \end{tabular}
   
 \end{center}
\caption{Distribution of distinct patterns per class used for training data creation and number of hits for each pattern. Last column shows no. of errors in N randomly sampled pattern instances.} \label{tab:lfs}
\end{table}

As our manually annotated dataset is too small to expect high accuracies for automatic prediction, we resort to data augmentation with weak supervision. Our approach proceeds as follows.

We first extract text segments from parliamentary debates from the German Bundestag (19th legislative term) and remove the debates in the test set from our unlabelled training corpus.
Each segment consists of a paragraph with multiple sentences, as annotated in the xml files. 
Please note that we do not assign labels to segments but to {\em instances of 1PL pronouns} in the segments.
We then apply a set of predefined patterns to identify instances of 1PL pronouns for each class in our annotation scheme. 
With the help of these patterns, we assign labels to the unlabelled training corpus and can now use this data to train a supervised ML system for pronoun disambiguation.
Below we explain the different steps in more detail.

\paragraph{Patterns} 
For pattern extraction, we make use of the spaCy DependencyMatcher which provides a flexible and efficient framework for defining search patterns over dependency trees.\footnote{See \url{https://spacy.io/api/dependencymatcher}. To generate the trees, we use the German \texttt{de\_core\_news\_sm} model also provided by spaCy.}

We combine the spacy DependencyMatcher with the Snorkel framework \cite{snorkel:2016,snorkel:2020}, a programmatic approach to data augmentation without manual labelling effort. Instead, Snorkel provides an API that allows users to write labelling functions that target specific labels in the annotation scheme. Those functions can consist of simple string matches but can also include more sophisticated features by including the predictions of pretrained classifiers or information from external knowledge bases.
While these labelling functions are expected to have low coverage and might also introduce a certain amount of noise, Snorkel addresses this problem by learning an unsupervised generative model over the output of the labelling functions, based on the (dis-)agreements between the predicted labels.
This approach is similar in spirit to previous work on quality estimation for annotations obtained from crowdsourcing \cite{hovy:etal:2013}.
The output of Snorkel is a set of probabilistic labels that can be used as input to any supervised ML classifier.

Table \ref{tab:lfs} shows the number of patterns used for each class and the number of hits, i.e., instances  extracted by each pattern from the unlabelled training data. Please note that the number of patterns is not very informative on its own, as patterns can make use of regular expressions, lemma lists and syntactic patterns over dependency trees, thus allowing us to extract a larger variety of diverse training examples than could be obtained based on simple string matches.

As an example, consider the following patterns used to extract labelled data for the {\sc Party} class. Our first pattern looks for instances of {\em wir, uns} (we, us) directly followed by a party name. This pattern can extract instances like \textit{Wir Gr\"{u}ne} or \textit{uns Liberale}. Another pattern looks for instances of {\em wir} as the subject of communication verbs like {\em kritisieren, hinterfragen} (criticize, question) etc., as those are usually statements refering to specific parties from the opposition.
A third example relies on future forms of {\em werden} (will) in combination with verbs of action, such as {\em schaffen, durchf\"{u}hren, investieren} (accomplish, execute, invest) to detect instances from the {\sc Government} class. This pattern would extract matches like \textit{wir werden Arbeitspl\"{a}tze schaffen} `we will create jobs' or \textit{Mindestens 2 Mrd. EUR werden wir in den sozialen Wohnungsbau investieren} `We will invest at least EUR 2 billion in social housing construction'.

The result of our pattern-based approach is a silver standard corpus with more than 36,000 labelled instances. To get an impression of the quality of the patterns, we randomly extracted 25 instances per class and manually inspected them  (last two columns in Table \ref{tab:lfs}). While most patterns seem to produce only a small amount of noise, some categories were more problematic. 
We found it particularly difficult to produce reliable patterns for {\sc People} and {\sc Generic} which is reflected in the low coverage and precision for the two classes (see \S \ref{sec:experiments}, Table \ref{tab:results-cv}).

\section{Experiments}
\label{sec:experiments}
\begin{table}[t]
 \begin{center}\small
  \begin{tabular}{|l|r|r|r|}
\hline 
{\bf wform} & {\bf class} & {\bf support} & {\bf DL} \\
\hline 
wir     & {\sc Parl}    & (185/600) & 9 \\
unser   & {\sc Country} &  (24/26)  & 2 \\
Wir     & {\sc Country} &  (65/240) & 9 \\
unserem & {\sc Country} &  (28/32)  & 4 \\
uns     & {\sc Country} &  (56/163) & 8 \\
unsere  & {\sc Country} &  (25/42)  & 6 \\
unserer & {\sc Country} &  (19/31)  & 7 \\
unseren & {\sc Country} &   (7/11)  & 4 \\
Uns     & {\sc Parl}    &   (1/2)   & 2 \\
Unser   & {\sc Country} &   (4/5)   & 2 \\
Unsere  & {\sc Country} &   (3/4)   & 2 \\
unseres & {\sc Country} &   (6/6)   & 1 \\
unsre   & {\sc Country} &   (1/1)   & 1 \\
Unsre   & {\sc Country} &   (2/2)   & 1 \\ 
 \hline
{\bf Total} &  & \multicolumn{2}{l|}{(426/1163)  Acc=36.6\%} \\
\hline
\end{tabular} 
\end{center}
\caption{Majority baseline, support and no. of distinct labels (DL) per pronoun word form in the test set.}\label{tab:majority}
\end{table}

We now explore the potential of our automatically created training set for disambiguating references of personal pronouns in political debates.
For that, we report results for three baselines and then present transfer learning experiments where we use our automatically created dataset for pretraining and then fine-tune the model on the manually created dataset.

\paragraph{B1: Majority Baseline} Our first baseline assigns each pronoun word form its most frequent label (Table \ref{tab:majority}). This results in an accuracy of 36.6\%. The last column shows the number of distinct labels (DL) per pronoun word form in the test set. The three most frequent word forms can occur with nearly any class ({\em Wir, wir}: 9 DL, {\em uns}: 8 DL), thus showing the difficulty of this task.

\paragraph{B2: Rule-based Baseline}  Our second baseline is a rule-based system that simply applies our pre-defined patterns to the testset and labels all matches with the respective labels. We use Snorkel's generative model (see \S \ref{sec:corpus}) for resolving ties between conflicting rules and report precision, recall and F1 for the rule-based approach. Table \ref{tab:results-baselines} (B2) shows that while we obtain a reasonable precision for some patterns ({\sc Country}: 92\%, {\sc Parl}: 91\%, {\sc Party}: 72\%), recall is a huge problem. For the two most difficult patterns, {\sc Generic} and {\sc  People}, we obtain not even one correct match.

\paragraph{B3: Feature-based Classification}  Our third baseline makes use of a conventional feature-based approach to text classification. For that, we consider the following features: (1) tf-idf ngram features (unigrams, bigrams, trigrams) for the left and right context of each 1PL pronoun, (2) the word form of the pronoun, and (3) named entities in the left and right context of the pronoun. We explored different settings for these features in a 5-fold cross-validation setup and observed best results for the feature values show in Table \ref{tab:feat}. We tested different classifiers (linear SVM, Ridge regression, SGD, decision trees, AdaBoost, Random Forests) and found that linear SVM gave us best results on our data (49.3\% acc.).\footnote{The models have been implemented with scikit-learn: \url{https://scikit-learn.org/stable/supervised_learning.html}.} 
Table \ref{tab:results-baselines} (B3) shows results for the linear SVM classifier. Results for other models and settings were in the range of 35-47\% acc.

\begin{table} 
\begin{center}
\begin{tabular}{|l|l|}
\hline
{\bf setting} & {\bf value} \\
\hline
left/right context size & 20 tokens \\
bow unigrams & yes \\
bow bigrams  & yes \\
bow trigrams & no \\
tfidf        & yes \\
lemmatisation & yes \\
stopwords    & no \\
feature selection & yes ($\chi^2$) \\
num features & 300 \\
NER in left/right context & no \\
\hline
\end{tabular}
\end{center}
\caption{Feature settings used for B3 (feature-based classification, Table \ref{tab:feat}).}\label{tab:feat}
\end{table}

\paragraph{Transfer Learning Model} Our model uses a simple transformer architecture, based on the sentence pair classifier implementation of Simpletransformers\footnote{\url{https://simpletransformers.ai}.} 
and the pretrained {\tt bert-base-german-dbmdz-cased} model.\footnote{The pretrained models are available from \url{https://github.com/dbmdz/berts}.} 
For details on parameter settings, please refer to Table \ref{tab:parameter} in the appendix.
The motivation behind modelling personal pronoun disambiguation as sentence pair classification is that we want to make the model aware of the pronoun's left and right context.
For that, we split each instance into two sequences where the first sequence encodes the left context of the pronoun in question and the second sequence includes the pronoun and its right context (see figure \ref{ex:split} below). Please note that our instances encode paragraphs, not sentences, and that S1 and S2 can thus include more than one sentences. In cases where the 1PL pronoun is positioned at the beginning of the paragraph, S1 will be empty.

\begin{figure}[h!]
\begin{center}
\begin{tabular}{|l|l|}
\hline
 Members of Congress , & \underline{\bf we} must work ... \\
 \hline 
 S1 & S2 \\
\hline
 \end{tabular}
 \end{center}
\caption{Setup for transfer learning using sentence pair classification; S1 encodes the left context of the 1PL pronoun, S2 the pronoun and its right context.} \label{ex:split}
 \end{figure}

\begin{table*}\small
 \begin{center}
  \begin{tabular}{|l|r|r|r|r|r|r|r|r|r|}
    \hline
            & \multicolumn{6}{c|}{\bf B2} & \multicolumn{3}{c|}{\bf B3}  \\
{\bf Class} & {\bf \#Gold} & {\bf \#Hits} & {\bf TP} &  {\bf Prec} & {\bf Rec} & {\bf F1} & {\bf Prec} & {\bf Rec} & {\bf F1}  \\
    \hline 
 BOARD      &   1 &  0 &  0 &  0 &   0 &  0 &  0 &  0 &  0     \\
 COUNTRY    & 411 & 37 & 34 & 92 &   8 & 15 & 53 & 72 & 61    \\  
 GENERIC    &  67 &  0 &  0 &  0 &   0 &  0 & 35 & 10 & 16    \\
 GOVERNMENT & 167 & 53 & 23 & 45 &  15 & 22 & 41 & 35 & 38    \\
 PARLIAMENT & 299 & 11 & 10 & 91 &   3 &  7 & 47 & 56 & 51   \\
 PARTY      & 103 & 17 & 13 & 76 &  13 & 22 & 49 & 30 & 37    \\
 PEOPLE     &  13 &  2 &  0 &  0 &   0 &  0 &  0 &  0 &  0   \\
 SPEC\_PERSON& 20 &  1 &  1 &  0 &   6 & 11 &  0 &  0 &  0    \\
 UNION      &  82 &  2 &  1 & 50 &   1 &  3 & 45 & 16 & 23    \\ 
 \hline
{\bf Total} &  1,163 & 123 &  83 & \multicolumn{3}{c|}{Acc = 7.0\%} & \multicolumn{3}{c|}{Acc = 49.3\%}  \\
\hline
\end{tabular}

\end{center}
\caption{Results for rule-based baseline (B2) and for the feature-based classification baseline (B3)
(precision, recall and f1 for individual classes and acc. for all instances).}\label{tab:results-baselines}
\end{table*}

\begin{table*}\small
 \begin{center}
  \begin{tabular}{|l|r|r|r|r|r|r|r|r|r|r|}
    \hline
            &  & \multicolumn{3}{c|}{\bf T1} & \multicolumn{3}{c|}{\bf T2} & \multicolumn{3}{c|}{\bf T3} \\
{\bf Class} & {\bf \#Gold} & {\bf Prec} & {\bf Rec} & {\bf F1} &  {\bf Prec} & {\bf Rec} & {\bf F1}  &  {\bf Prec} & {\bf Rec} & {\bf F1} \\
    \hline
 BOARD      &   1 &      0 &  0 &  0  &   0 &  0 &  0  &  0  &  0 &  0 \\
 COUNTRY    & 411 &     58 & 65 & 62  &  65 & 63 & 64  & 56  & 66 & 60 \\  
 GENERIC    &  67 &     29 & 13 & 18  &  20 & 16 & 18  & 50  &  7 & 13 \\
 GOVERNMENT & 167 &     40 & 36 & 38  &  40 & 47 & 43  & 40  &  4 &  7 \\
 PARLIAMENT & 299 &     50 & 64 & 56  &  56 & 54 & 55  & 45  & 78 & 57 \\
 PARTY      & 103 &     56 & 36 & 44  &  52 & 54 & 53  & 43  & 32 & 36 \\ 
 PEOPLE     &  13 &      0 &  0 &  0  &   9 & 23 & 13  &  0  &  0 &  0 \\ 
 SPEC\_PERSON& 20 &     17 & 10 & 12  &   6 &  1 &  8  &  0  &  0 &  0 \\  
 UNION      &  82 &     28 & 17 & 21  &  36 & 24 & 29  & 60  &  9 & 16 \\  
 \hline
{\bf Total} &  1,163 &  \multicolumn{3}{c|}{Acc = 50.2\%} & \multicolumn{3}{c|}{Acc = 50.9\%} & \multicolumn{3}{c|}{Acc = 51.8\%} \\
\hline
\end{tabular}

\end{center}
\caption{Results for 5-fold cross-validation for 3 transfer learning settings. T1: training on testset only; T2: training on testset + augmented data; T3: pretraining on augmented data and fine-tuning on testset (precision, recall and f1 for individual classes and acc. for all instances).}\label{tab:results-cv}
\end{table*}

\paragraph{Results for 5-fold cross-validation} We now report cross-validation results on our small, manually annotated dataset (Table \ref{tab:results-cv}). As we do not have enough data to create a representative validation set for model selection, we report preliminary results for all models (T1, T2, T3) after 25 epochs of training. This procedure has to be taken with a grain of salt and will be addressed, once we have more annotated data.

The results show that even a small number of annotated instances yields substantial improvements over the majority baseline (Table \ref{tab:majority}) and accuracy increases from 36.6\% to over 50\%. 
The results, however, are only slightly higher than the ones for the SVM (Table \ref{tab:results-baselines}, B2). 
Table \ref{tab:results-cv}, (T2) shows results for merging the hand-annotated data with the noisy labels. In order not to outweigh the manual annotations, we downsampled the additional training data to at most 300 new instances per class.  
This setting results in only minor improvements (from 50.2 to 50.9\% acc.). In our third setting, we use the noisy labels for an additional pretraining step before fine-tuning the model on the hand-annotated data. This yields another small improvement and increases accuracy to 51.8\%.

\paragraph{Discussion}

The somewhat disappointing results for our data augmentation strategy might have several reasons.
First, it is conceivable that we need to put more effort into creating a) more precise and b) more diverse rules, and c) to improve coverage.
Results on a held-out dataset, created by the same rule-based approach, show that our model is perfectly able to learn the annotations in the weakly supervised data, achieving an accuracy of 97.6\% on the held-out data. This shows that despite our efforts to minimise lexical cues and rely more on syntactic patterns, our augmented training data is highly biased and does not enable the model to learn good generalisations for each class.

While improving coverage for the rule-based approach might ameliorate the problem, it is also possible that the pattern-based approach is more suitable for less ambiguous classification tasks, such as spam detection or offensive language detection, where we only have a small number of classes that are more clearly divided and where it is easier to create patterns with a high precision and coverage.

\section{Conclusions}
\label{sec:conclusions}

In the paper, we investigated what kinds of collectives 1PL pronouns refer to in parliamentary debates. To this end, we developed an annotation scheme that assigned references to one of nine categories and explored how well human annotators agree when assigning those categories. Our annotation study showed a substantial agreement of $>0.8 \alpha$ between two human raters. 
We then presented a preliminary ana\-lysis of the use of 1PL pronouns as a rhetorical device and pointed to some crucial differences between the parties as well as between members of the government and opposition parties. We subsequently explored how well we are able to automatically resolve ambiguous 1PL pronouns in parliamentary debates, using transfer learning and data augmentation. While our preliminary results are promising, there is  room for improvment before we can apply our work to large-scale analysis of pronoun references in political text.

In future work, we plan to improve the accuracy of 1PL pronoun resolution by creating more training data, but also by improving the model itself. Possible ways to do so include providing the model with more information on the speaker, such as the speaker's name, party affiliation or whether or not the speaker is part of the government. Other improvements might come from jointly modelling 1PL pronouns in context, instead of looking at them one at a time.

\section*{Acknowledgments}

This work was supported in part by the SFB 884 on the Political Economy of Reforms at the University of Mannheim (projects B6 and C4), funded by the German Research Foundation (DFG).

\bibliography{refs}
\bibliographystyle{acl_natbib}

\appendix

\section{Appendices}
\label{sec:appendix}

\begin{table*}[t]
\begin{center}\small
\begin{tabular}{ | l | l | l |}
\hline
\thead{Category} & \thead{Description} & \thead{Examples} \\
\hline
{\sc COUNTRY} & \makecell[l]{
Refers to the country as a\\
geo-political unit or to all \\
citizens of this country. \\
\\
TEST: can be replaced by \\
\tabitem "we Germans" \\
\tabitem "our country" \\
\tabitem "the German X" \\
}
& \makecell{
{\bf Wir} haben 2 Weltkriege verloren. \\
{\bf Wir} sind Exportweltmeister. \\
{\bf Wir} sind Papst. \\
{\bf Wir} als nationale Schicksalsgemeinschaft. \\
{\bf Wir} dürfen uns nicht vom Rest der Welt abschotten. \\
{\bf Unser} Grundgesetz / unsere Demokratie}. \\
\hline

\pbox{2cm}{PEOPLE} &
\makecell[l]{
Refers to a (possibly large) group\\
of people that are not defined\\
by their nationality but by shared\\
social variables or characteristics \\
such as age, gender, class, religion, \\
profession, ... \\
Also used for references to society\\
that are not limited to Germany as \\
a geo-political unit.} &  
\makecell{
wie {\bf wir} Christen uns verhalten \\
{\bf Wir} als arbeitende Bevölkerung \\
{\bf Wir} Älteren, {\bf Wir} Rentner \\
{\bf Wir} Steuerzahler, {\bf Wir} Pendler} \\
\hline
 
\pbox{2cm}{PARTY} & 
\makecell[l]{
Refers to members of a  \\
specific party (including \\
coalitions of like-minded \\
parties, e.g., on the \\
supranational level)} &
\makecell{
{\bf Unser} Antrag geht einen entscheidenden Schritt... \\
{\bf Wir} werden diese Regierung weiter kritisieren. \\
{\bf Wir} Liberale haben schon vor Jahren gesagt, ...} \\
\hline

\pbox{2cm}{PARL} &
\makecell[l]{
Refers to all members of the \\
parliament (also references \\
to both, government and \\
opposition)} &
\makecell{
{\bf Wir} Abgeordnete sind vom Volk gewählt. \\
In diesem Haus debattieren {\bf wir} heute... \\
Lassen Sie {\bf uns} diesen Antrag heute beschließen.} \\
\hline

\pbox{2cm}{GOVERN} &
\makecell[l]{
Refers to all members of \\
the government} &
\makecell{
{\bf Wir} haben entscheidende Schritte getan,\\
um die Digitalisierung zu fördern. \\
{\bf Wir} haben Familien entlastet und \\
die Arbeitslosigkeit bekämpft} \\
\hline

\pbox{2cm}{UNION} & 
\makecell[l]{
Refers to geo-political \\
groups on a supranational \\
level, e.g., the EU, the NATO,\\
etc.} &
\makecell{
{\bf Wir} in der EU müssen zusammen einen Weg\\
finden, wie wir unsere Sicherheitspolitik gestalten.} \\
\hline

\makecell[l]{SPEC\_PERS \\
(GROUPS)} & 
\makecell[l]{
Refers to groups of specific \\
individuals or members of  \\
more than one group} & 
\makecell{
Sie haben die PKK und die YPG in einen Topf\\
geworfen, {\bf wir} sind aber nicht deckungsgleich.\\
Frau Merkel und ich, {\bf wir} haben darüber \\
lange diskutiert. \\
{\bf Wir}, die deutsche und die israelische Regierung}\\
\hline

GENERIC &
\makecell[l]{ 
Generic uses of {\em we/us} that can be  \\
replaced by {\em one/you} (German: man/es\\
gibt) or {\em unser/e} can be replaced by {\em diese}. \\
We assume a generic reading if {\em we/us}\\
refers to the whole world/universe. \\
}
&
\makecell{
Das brauchen {\bf wir} überall in der Welt \\
$\rightarrow$ das braucht man überall... \\
In den letzten Jahren haben {\bf wir} viel \\
über den Wandel der Gesellschaft gehört \\
$\rightarrow$ hat man viel gehört über... \\
Woran {\bf wir} uns noch in 100 Jahren erinnern werden\\ 
$\rightarrow$ Woran man sich noch in...\\
die schwierigen Probleme {\bf unserer} Zeit \\
$\rightarrow$ dieser Zeit \\
In einer Welt, in der {\bf wir} über 222 gewaltsam \\
ausgetragene Konflikte haben\\
$\rightarrow$ in der es ... gibt} \\
\hline

BOARD &
\makecell[l]{ 
Refers to members of a \\
board / commission / committee \\
/political organisation on the\\
subnational level (subgroups of \\
the parliament/government)} &
\makecell{ 
{\bf Wir} haben im Untersuchungsausschuss \\
viel diskutiert...\\
Im Coronakabinett haben {\bf wir} beschlossen...\\
Im Agrarausschuss haben {\bf wir} ...} \\

\hline
\end{tabular}
 \end{center}
\caption{Overview of the annotation scheme for 1PL references in parliamentary debates.}\label{tab:schema-long}
\end{table*}

\begin{table*}
 \begin{center}\small
  \begin{tabular}{|l|r|r|r|r|r|r|r|r|r|}
\hline
\backslashbox{A2}{A1}& {\bf BOARD} & {\bf COUNTRY} & {\bf GENERIC} & {\bf GOVREN} & {\bf PARL} & {\bf PARTY} & {\bf PEOPLE} & {\bf SPECPER} &  {\bf UNION} \\
\hline
{\bf BOARD}   & 0 &   0 &  0 &   0 &   0 &  0 &  0 &  0 &  0 \\
\hline
{\bf COUNTRY} & 0 & 385 &  8 &   4 &  14 &  3 &  1 &  3 & 12 \\
\hline
{\bf GENERIC} & 0 &   4 & 46 &   1 &  13 &  0 &  2 &  0 &  1 \\
\hline
{\bf GOVERN}  & 0 &   7 &  1 & 146 &   8 &  7 &  0 &  1 &  4 \\
\hline
{\bf PARL}    & 1 &   8 & 14 &   2 & 248 &  0 &  2 &  4 &  4 \\
\hline
{\bf PARTY}   & 0 &   1 &  0 &   2 &   5 & 96 &  0 &  0 &  1 \\
\hline
{\bf PEOPLE}  & 0 &   1 &  2 &   0 &   2 &  0 & 11 &  0 &  0 \\
\hline
{\bf SPECPER}   & 0 &   0 &  0 &   0 &   0 &  0 &  1 & 10 &  0 \\
\hline
{\bf UNION}   & 0 &   1 &  2 &   4 &   0 &  1 &  0 &  2 & 61 \\
\hline
\end{tabular}
 \end{center}
\caption{Confusion matrix for the manual resolution of referents of ambiguous pronouns in parliamentary debates (A1: Annotator 1, A2: Annotator 2).}\label{tab:cm}
\end{table*}

\begin{table*}
 \begin{center}\small
  \begin{tabular}{|l|r|}
\hline
{\bf Name} & {\bf Value} \\
\hline
  attention\_probs\_dropout\_prob &  0.1 \\
  hidden\_act &  gelu \\
  hidden\_dropout\_prob &  0.1 \\
  hidden\_size &  768 \\
  layer\_norm\_eps &  1e-12 \\
  max\_position\_embeddings &  512 \\
  num\_attention\_heads &  12 \\
  num\_hidden\_layers &  12 \\
  transformers\_version &  4.6.1 \\
  type\_vocab\_size &  2  \\
  vocab\_size &  31102  \\
\hline
\end{tabular}
 \end{center}
\caption{Parameters/settings used in our experiments.}\label{tab:parameter}
\end{table*}

\end{document}